\documentclass{article}
\usepackage{ijcai19}

\usepackage{times}
\usepackage{soul}
\usepackage{url}
\usepackage[hidelinks]{hyperref}
\usepackage[utf8]{inputenc}
\usepackage[small]{caption}
\usepackage{graphicx}
\usepackage{amsmath}
\usepackage{booktabs}
\usepackage{algorithm}
\usepackage{latexsym}
\usepackage{subfig}
\usepackage{xcolor}
\usepackage{color}
\usepackage{xspace}

\usepackage{multirow}
\usepackage{multicol}
\usepackage[noend]{algpseudocode}
\usepackage{tabularx}
\usepackage{CJKutf8}
\usepackage{paralist}
\newcommand{\tabincell}[2]{
\begin{tabular}{@{}#1@{}}#2\end{tabular}
} 
\newcommand \newcite[1]{{ \citeauthor{#1}~\shortcite{#1} }}

\newcommand{\method}{CAMIT\xspace}

\title{Correct-and-Memorize: Learning to Translate from Interactive Revisions}

\author{
Rongxiang Weng$^{1,2}$\thanks{Equal contribution, part of this work was done while Rongxiang Weng was a research intern at ByteDance AI Lab. }\and
Hao Zhou$^{3*}$\and
Shujian Huang$^{1,2}$\thanks{Corresponding author.} \and
Lei Li$^3$\and
Yifan Xia$^{1,2}$\And
Jiajun Chen$^{1,2}$
\affiliations
$^1$National Key Laboratory for Novel Software Technology, Nanjing, China\\
$^2$Nanjing University, Nanjing, China\\
$^3$ByteDance AI Lab, Beijing, China
\emails
wengrx@nlp.nju.edu.cn, zhouhao.nlp@bytedance.com,
huangsj@nlp.nju.edu.cn,
lileilab@bytedance.com,
\{xiayf, chenjj\}@nlp.nju.edu.cn
}
\hypersetup{draft}
\begin{document}

\maketitle

\begin{abstract}

State-of-the-art machine translation models are still not on par with human translators. 
Previous work takes human interactions into the neural machine translation process to obtain improved results in target languages. 
However, not all model-translation errors are equal -- some are critical while others are minor. 
In the meanwhile, the same translation mistakes occur repeatedly in a similar context. 
To solve both issues, we propose \method, a novel method for translating in an interactive environment. 
Our proposed method works with critical revision instructions, therefore allows human to correct arbitrary words in model-translated sentences. 
In addition, \method learns from and softly memorizes revision actions based on the context, alleviating the issue of repeating mistakes. 
Experiments in both ideal and real interactive translation settings demonstrate that our proposed \method enhances machine translation results significantly while requires fewer revision instructions from human compared to previous methods\footnote{Source code is available at: https://github.com/wengrx/CAMIT}.


\end{abstract}

\section{Introduction}
\begin{figure}[ht]
\centering
\subfloat[The previous uni-directional interactive NMT method could fix the mistake of ``trade'' $\rightarrow$ ``free trade'' automatically after revising ``to discuss'' $\rightarrow$ `` discuss''.]{
 \label{fig:Left2right}
\includegraphics[scale=.58]{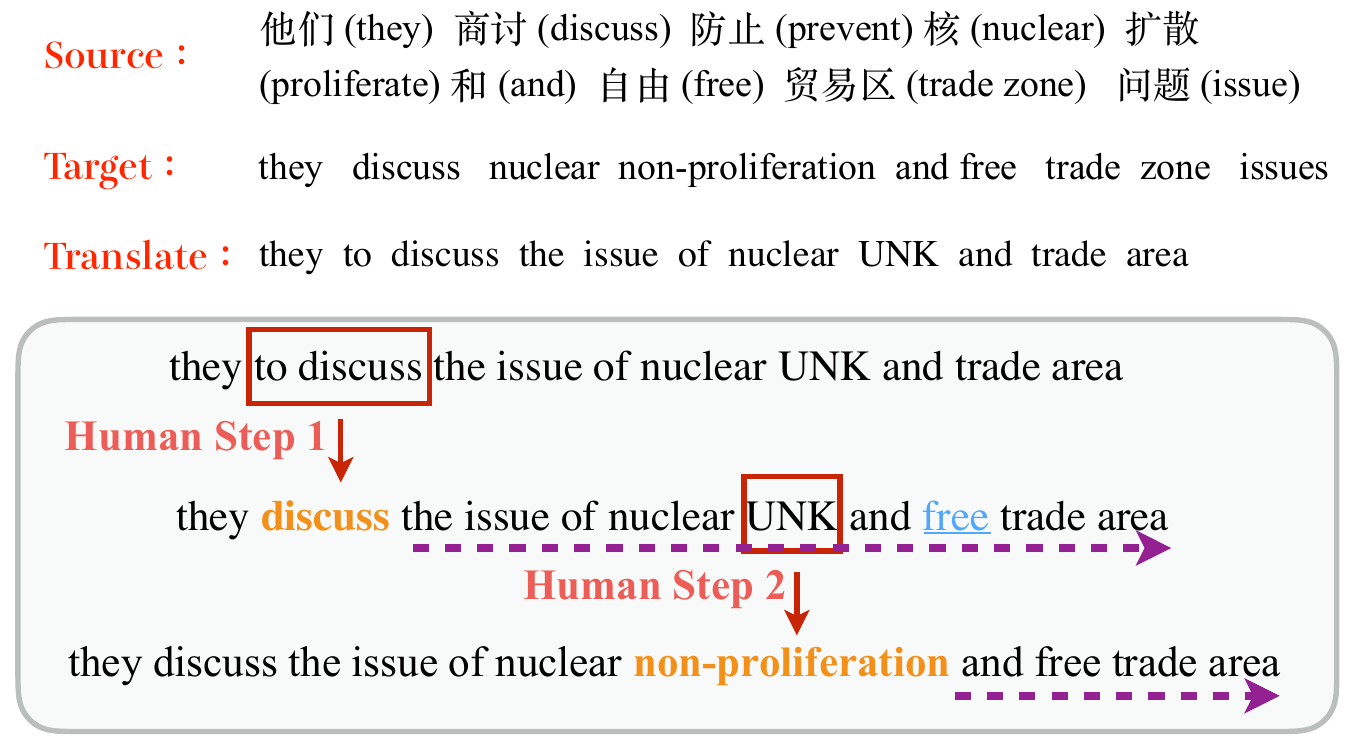}}\\
\subfloat[Our proposed \method takes the most critical revision ``\emph{UNK}'' $\rightarrow$ ``non-proliferation'' and automatically fixes the rest mistakes across the whole sentence.]{
 \label{fig:bidirect}
\includegraphics[scale=.58]{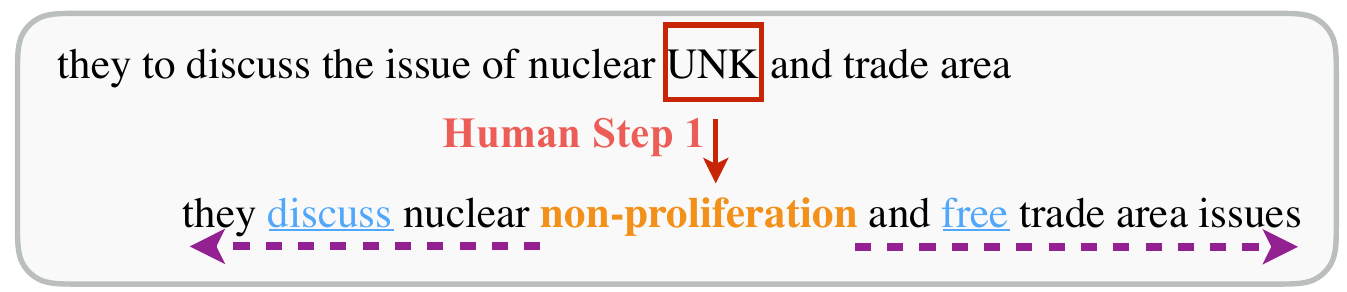}}
\label{fig:intro_example}
\caption{Illustration of the previous and our proposed interactive NMT methods. Words with bold and orange fonts are revised by human.  Those with underlines and blue fonts are automatically fixed by models. Note our proposed \method only needs one revision while the previous method requires two. }
\end{figure}


Recently, sequence-to-sequence models~\cite{sutskever2014sequence,Cho2014Learning} have gain superior performance in machine translation~\cite{Bahdanau2015Neural,vaswani2017attention}.
Yet these state-of-the-art neural machine translation~(NMT) models 
still fail to generation target sentences with comparable quality as human translators. 

To obtain a high-quality translation, there are a few recent works attempt to incorporate human revision instructions into the algorithmic translation process. 
For example, previous interactive NMT~\cite{sanchis2014interactive,Alvaro2016Interactive,knowles2016neural} proposes to ask human to revise the translation output from the beginning of a sentence to the end (i.e. from the left to right), and regenerates the partial translation on the right side of the revised token. We refer to these models as \textit{uni-directional} interactive models. 
The interactive steps can repeat multiple times until a satisfactory sentence is obtained.
In this case, given human revisions for model-translation mistakes, the process could potentially fix some minor mistakes automatically, exploiting the base translation models' ability to save human efforts~(See Figure~\ref{fig:Left2right}).


In this paper, we make a few observations about neural machine translation and note two issues in previous interactive NMT approaches. 
First, the previous uni-directional interactive NMT method could only possibly fix those errors to the right of human-revised words. 
\newcite{cheng2016a} show that revising critical mistakes first could significantly reduce the number of revisions. 
However, their intuition could not be directly applied to current uni-directional NMT models, because after revising critical mistakes, uni-directional interactive models cannot correct minor mistakes to the left of the revision, even with advanced decoding algorithm~\cite{hokamp2017lexically,post2018fast,hasler2018neural}.
Since the uni-directional interactive model only regenerates the partial sentence to the right of revisions, instead of updating the whole sentence.
This leads to the constraint that each human revision should be proposed at the left most errors. 
Thus, human annotators have to revise these remaining mistakes manually, which is not efficient enough.

Second, neural machine translation models often keep making the same mistakes with similar discourse and domain context. 
Previous interactive NMT models rarely exploit the revision history to avoid similar mistakes.
Our basic intuition is that previous sentences which have been corrected by human may be of greatly help to the translation of the following sentences when they are in the same discourse or domain. 
Learning from past revisions could preventing NMT models from making similar mistakes.

To address the above issues, we propose a correct-and-memorize framework for interactive machine translation~(\method), a novel framework to perform efficient corrections and then softly memorize those corrections for enhancing translation results. 

First of all, \method contains an improved decoder given a sentence and one revised word. 
\method performs the interactive NMT from both ends of the sentence, updating the whole sentence after getting a revision.
Different from previous work about bi-directional decoder~\cite{Liu2016Agreement,Mou2016sequence,liu2018bfgan}, our method is designed to take human revisions for interactive NMT.  
In such case, human can revise the most critical mistake first in an arbitrary position of the sentence; and after that the model will update the whole sentence, fixing minor mistakes automatically.
Thus the efficiency of interaction could be significantly improved~(Figure \ref{fig:bidirect}).

Secondly, we propose to learn from and memorize interactive revisions to enhance the translation results. 
The revision history could be learned in the \textit{word level}  and the \textit{sentence level}, respectively.
At the word level, we propose a key-value memory mechanism~\cite{westion2014memory} called \textit{revision memory}, to remember prior revisions in the interactive process. 
At the sentence level, we use the \textit{online learning} approach to fine-tune our base translation model on prior revised sentences,  adapting our model parameters to the specific discourse or domain.
When translating a new sentence, our model avoids generating similar mistakes in the past by looking up the revision memory, which is especially helpful to the translation of rare words. 

Experiments show that our proposed \method outperforms the previous interactive methods in both ideal and real machine translation settings.
Our method's decoder obtains the improvement of 17 BLEU points with merely two revisions in the ideal and 8 BLEU points with 1.81 revisions in the real environments, respectively.
Using interaction history further helps improve interactive efficiency and translation quality.


\section{Background}
Neural machine translation~(NMT) is based on a standard Seq2Seq model, which adopts an encoder-decoder architecture for sentence modeling and generation~\cite{sutskever2014sequence,Cho2014Learning,Bahdanau2015Neural,vaswani2017attention}.
The encoder summarizes the source sentence into an intermediate representation, and the decoder generates the target sentence from left to right.

Formally, let ${\mathbf x}=\{x_1, \cdots, x_i, \cdots\}$ be a given input sentence and ${\mathbf y}=\{y_1,\cdots, y_j,\cdots\}$ be the output sentence.
The encoder transforms the sentence to a sequence of annotations $\textbf{H}$, with each $\mathbf{h}_i$ being the annotation of input word $x_i$.

Based on the source annotations, the decoder generates the translation by predicting a target word $y_j$ at each time step $j$:
\begin{equation}
P(y_j|y_{<j},\mathrm{\mathbf{x}}) =\mathrm{softmax}(g(\mathbf{s}_j)), \label{eq: prob}
\end{equation}
where $g$ is a non-linear activation function, and $\mathbf{s}_j$ is the decoding state for time step $j$, computed by
\begin{equation}
\mathbf{s}_j=\overrightarrow{f}(y_{<j}, \mathbf{c}_j) \label{s}.
\end{equation}
Here $\overrightarrow{f}$ is a recurrent unit~\cite{Cho2014Learning} or self-attention network~\cite{vaswani2017attention}.
$\mathbf{c}_j$ is a vector summarizing relevant source information. It is computed by the attention mechanism~\cite{Luong2015Effective,vaswani2017attention}.
\begin{equation}
\mathbf{c}_j = \text{ATT}(\textbf{r},\textbf{H}),
\label{eqn-context}
\end{equation}
where the $\textbf{r}$ is a state from last time step or previous layer when the $\overrightarrow{f}$ is a recurrent unit or a self-attention network, respectively.

In the training stage, $\textbf{y}$ is the gold sentence provided by training set. NMT models optimize the networks by maximizing the likelihood denoted by $\mathcal{L}_{L}$.
\begin{equation}
\mathcal{L}_{L}=\frac{1}{|\textbf{y}|}\sum_{j=1}^{|\textbf{y}|}\log P(y_{j}|y_{<j},\textbf{x}). \label{loss}
\end{equation}
where $|\textbf{y}|$ is the length of $\textbf{y}$ and $P(y_{j}|y_{<j},\textbf{x})$ is defined in Equation~\ref{eq: prob}.

\section{The Proposed \method}

Practically, there may be multiple mistakes in the translation output.
Prior work propose to incorporate human efforts to boost the translation performance from left to right, both in the statistical machine translation (SMT)~\cite{barrachina2009statistical,gonzalez2013interactive} and the neural machine translation (NMT)~\cite{knowles2016neural,Alvaro2016Interactive,wuebker2016models}.
However, current interactive model is not efficient enough in both decoding and learning, as previously discussed.

\begin{figure}[t]
    \centering
    \includegraphics[scale = 0.30]{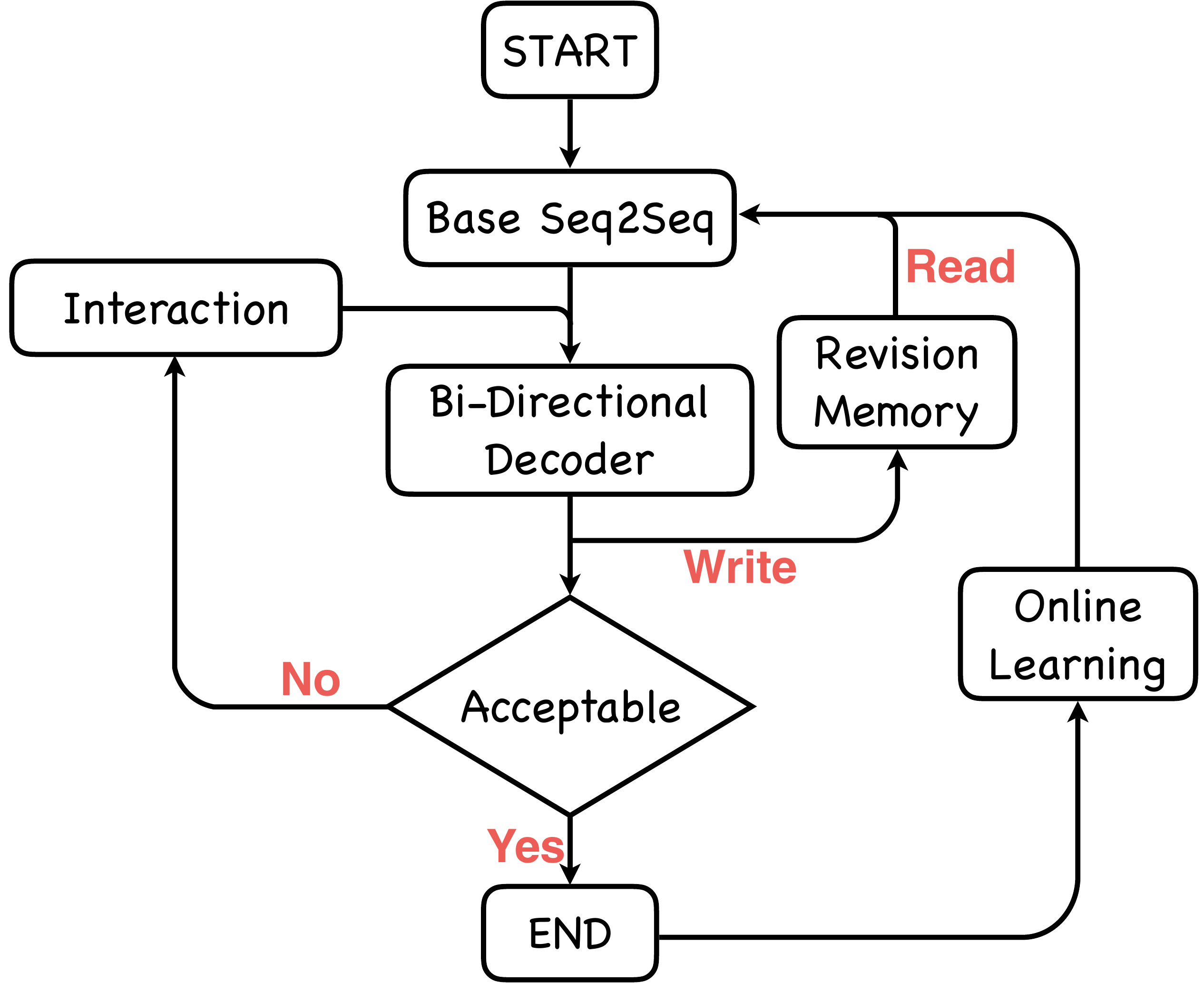}
    \caption{\label{fig:flowchart} The procedure of our proposed interactive model.}
\end{figure}

In this section, we propose \method, to improve the interactive efficiency in these two aspects.
As shown in Figure \ref{fig:flowchart}, after each interaction from human, we propose a \textit{sequential bi-directional decoder} for updating the whole sentence.
All revisions are written into a \textit{revision memory}, and the base Seq2Seq model will read from it latter for avoiding making same mistakes.
Additionally, previously revised sentence will be learned by an \textit{online learning} approach.

\subsection{Preliminary}
\label{sec:preliminary}
The interactive process always contains many sentences; and each sentence may need to be revised for many times. 
One \textit{revision} is the process that human translator manually corrects one mistake in a sentence. 

Commonly, the revision can include several operations: \textit{replacement}, \textit{insertion} and  \textit{deletion}. 
Following previous work~\cite{Alvaro2016Interactive,knowles2016neural,cheng2016a}, we focus on the replacement and others can be implemented with several replacement operations.
For example, deletion could be converted to replace incorrect words with spacing, and the insertion could be converted to replace spacing with inserted words.
Furthermore, we only apply the word level revision in this paper, other types of revision, such as phrase, subword, etc., can also be used in our interactive model.

Because human may need to give multiple revisions to one sentence, we refer to the multiple revisions for one sentence as one \textit{round} of interaction.
Commonly, adjacent sentences in one interactive process, which are usually in the same discourse or domain, are related.
We model the interactive process of many related sentences as one \textit{session}.
One session may contain many rounds, which could be accomplished by one or more human translators.

\subsection{Sequential Bi-Directional Decoding}
\label{sec:bi-directional}

In the interactive process, let ${\mathbf y}=\{y_1, \cdots, y_j, \cdots\}$ be the initial sequence of words, output by the base Seq2Seq model. If human translator revises the word $y_j$, the output is divided into 2 parts:
\begin{align}
    \{\underbrace{\cdots,y_{j-2},y_{j-1}}_{\text{left}}, y_j, \underbrace{y_j,y_{j+1},\cdots}_{\text{right}}\}. \nonumber
\end{align}
Uni-directional interactive model only updates the \textit{right} part  ~\cite{Alvaro2016Interactive,knowles2016neural}, ignoring potential mistakes in the \textit{left} part, which means the revised word should always be the left most error, otherwise the errors in the left part will never be corrected.

We purpose a sequential \textit{bi-directional decoder} for interactive NMT, which can update both parts of the sentence.
The proposed model includes two decoders: a forward decoder $\overrightarrow{f}$ and a backward decoder $\overleftarrow{f}$.
$\overrightarrow{f}$ and $\overleftarrow{f}$ share the same encoder, but working sequentially.

Formally, in the interactive process, the forward decoder $\overrightarrow{f}$ first acts as the basic model to generate the basic translation:
$\textbf{y} = \{y_{1},\cdots, y_j, \cdots\}$. 
Given a \textit{revision} $\{ y_j \rightarrow y^{r}_j\}$, i.e. revising $ y_j $ to $ y^{r}_j$, $\overrightarrow{f}$ then generates the \textit{new right} part for the translation, the same as in the uni-directional model. 
The new decoding state $\mathbf{s}'_{j+1}$ for $\overrightarrow{f}$ at the position $j+1$ is computed as: 
\begin{equation}
\overrightarrow{\mathbf{s}}'_{j+1}=\overrightarrow{f}(y^{r}_{j},\mathbf{c}'_{j+1}),
\end{equation}
and the new sentence is
\begin{align}
    \{\cdots,y_{j-2},y_{j-1}, y^{r}_j, \underbrace{y'_{j+1},y'_{j+2},\cdots}_{\text{new right}}\}. \nonumber
\end{align}
Here words before  $y^{r}_{j}$ are kept the same, the probabilities of words in the new right part is computed the same as in Equation \ref{eq: prob}.



Different from previous approaches, we then feed the inverted new right part
$\{\cdots,y'_{j+2},y'_{j+1},y^{r}_{j}\}$ into  $\overleftarrow{f}$, obtaining a new decoding state at the current revising position:
\begin{align}
\overleftarrow{\textbf{s}}'_{j-1} &= \overleftarrow{f}(y^{r}_{j},\textbf{c}'_{j-1}).
\end{align}
Then starting from $\textbf{s}'_{j-1}$, $\overleftarrow{f}$ generates the \textit{new left} part $\{y'_{j-1}, y'_{j-2},\cdots\}$ according to a reversed decoding probability:
\begin{equation}
 P(y'_{j-1}|y'_{>j},y^{r}_{j},\textbf{x})=\text{softmax}(g(\textbf{s}'_{j-1}).
\end{equation}
Finally, the output is
\begin{align}
\{ \underbrace{\cdots, y'_{j-2}, y'_{j-1}}_{\text{new left}}, y^{r}_{j}, \underbrace{y'_{j+1},y'_{j+2},\cdots}_{\text{new right}} \}.  \nonumber
\end{align}
Note that, the length of the new left part may be not equal to the original left part.

In this way, the whole sentence is updated jointly by the two decoders.
With the human revision, the new right part is expected to be better than the right part. Based on this better right part, the new left part is expected to become better as well.

The training stage is similar with multi-task models~\cite{Dong2015Multi,Luong2015Multi},  both decoders could be trained using cross-entropy as the objective:
\begin{align}
\mathcal{L} &= \mathcal{L}_{L} + \mathcal{L}_{R}, 
\end{align}
where $\mathcal{L}_{R}$ is computed by:
\begin{align}
\mathcal{L}_{R} &=\frac{1}{|\textbf{y}|}\sum_{j=1}^{|\textbf{y}|}\log P(y_{j}|y_{>j},\textbf{x}).
\end{align}
The $P(y_{j}|y_{>j},\textbf{x})$ and $\mathcal{L}_{L}$ are defined in Equation \ref{eq: prob} and Equation \ref{loss}, respectively. $|\textbf{y}|$ is the length of $\textbf{y}$.

We have shown how to update the sentence after a single \textit{revision}. However, another challenge for interactive NMT happens when human translator performs several revisions in one \textit{round}, because the interactive process should regenerate the translation with all the revisions considered. 

To solve the problem, we propose to combine the grid beam search~\cite{hokamp2017lexically} with our bi-directional decoder. 
Intuitively, the grid beam search adopts a grid to store the partial translations that containing the previous revisions. 
After the gird search, all previous revisions will be included in the final translation output.
We propose to perform the grid beam search twice in our bi-directional decoding framework. 
\begin{figure}[t]
\centering
\includegraphics[scale = 0.31]{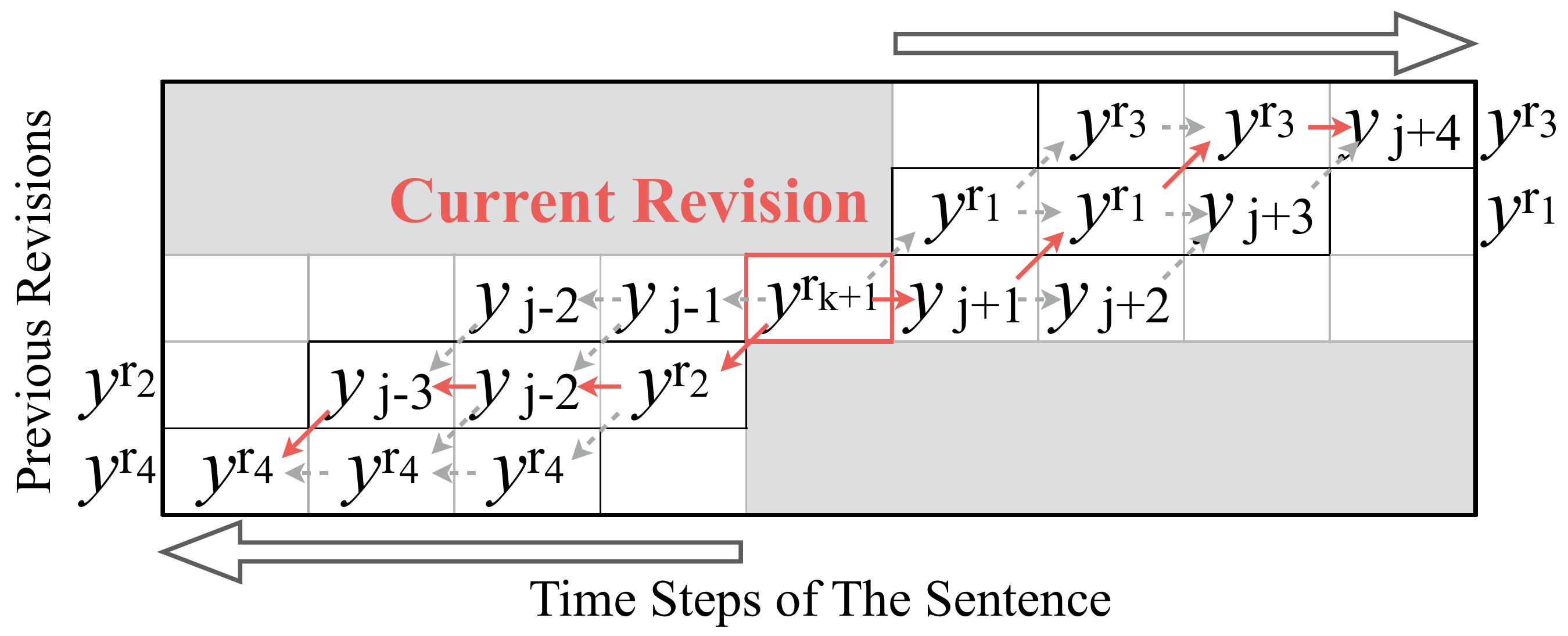}
\caption{\label{fig:grid serach} The restricted bi-directional grid beam search.} 
\end{figure}

Specifically, given a current revision $y_{k+1}^r$ and several previous revisions~$\{ y^r_{1}, y^r_{2}, y^r_{3}, y^r_{4} \}$, in which $\{y^r_{1}, y^r_{3}\}$ are in the right part of the current revision and $\{y^r_{2}, y^r_{4}\}$ are in the left part of the current revision, respectively.
We will first perform the gird beam search starting from  $y_{k+1}^r$ with the forward decoder $\overrightarrow{f}$, obtaining the new right part containing previous revisions~($\{y^r_{1}, y^r_{3}\}$) right  to $y_{k+1}^r$.
Then given the new right part, we will perform the grid beam search the second time from right to left using the backward decoder $\overleftarrow{f}$, obtaining the new right part containing previous revisions~($\{y^r_{2}, y^r_{4}\}$) left to $y_{k+1}^r$.
In such case, all previous revisions~($\{y^r_{2}, y^r_{4}, y^r_{1}, y^r_{3}\}$) are included in the final output, keeping their sequential order.
Figure \ref{fig:grid serach} gives an illustration of the bi-directional grid search process.

\subsection{Learning from Interaction History}
The interaction history is very helpful to improve the efficiency of interactive models.
In this section, we propose methods to learn from the interactive history by online learning and adopting revision memory.


\subsubsection{Revision Memory}


We first propose to exploit the interaction history in the word level, using a key-value memory~\cite{westion2014memory}, referred to as the \textit{revision memory}, to memorize previously performed revisions of the session.
This makes sense because some specific rare words, e.g. mostly out-of-vocabulary words~(\textit{OOV}), occur multiple times in a discourse. The proposed mechanism can help avoid making the same mistakes by memorizing and copying past revisions of the current session.

\begin{figure}[t]
    \centering
    \includegraphics[scale = 0.35]{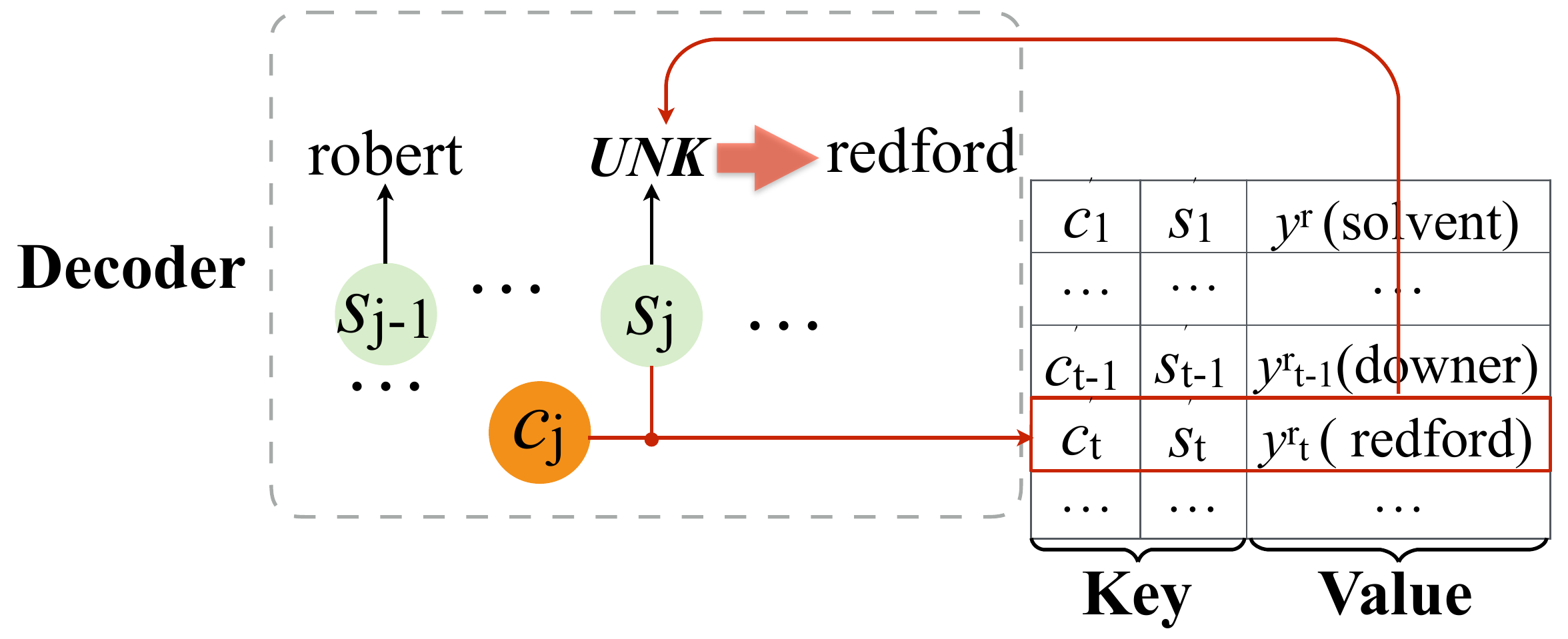}
    \caption{\label{fig: hrm} Using revision memory in the decoding stage.} 
    \end{figure}

Specifically, as shown in Figure \ref{fig: hrm}, the revision memory
\begin{equation}
\mathbf{M}=\{(<\textbf{s}'_{1},\textbf{c}'_{1}>,y^{r}_{1}),\cdots,(<\textbf{s}'_{T},\textbf{c}'_{T}>,y^{r}_{T})\} \nonumber
\end{equation}
 consists of many items, each of which has a key (\textit{revision context}) and a value~(\textit{revised word}).
The revised word is used to denote the revision itself. The revision context is used to determine whether the memorized revision should be performed again at the current decoding step. 
We define the revision context to include two parts: the decoding state $\textbf{s}'$~(Equation \ref{s}) and the context vector $\textbf{c}'$~(Equation \ref{eqn-context}) of the past revision, which represent the revision context in aspects of the language model and the translation model, respectively.

Every revision of the current session will be \textit{written} into the memory. When generating a word in decoding, the model will \textit{read} the revision memory, trying to automatically fix mistakes with a \textit{copy} mechanism~\cite{gu2016incorporating}.
The final output distribution of word $w$ is computed by distributions from the decoder and from the revision memory: 
\begin{align}
P_{j}(w)&= (1-\theta_{j})*P_{j}(w)+\theta_{j}*r_{j}(w), \\
r_{j,t} &= \text{softmax}(W_{1}|\textbf{s}_{j}\cdot\textbf{s}'_{t}| + W_{2}|\textbf{c}_{j}\cdot\textbf{c}'_{t}|), 
\end{align}
where $r_{j,t}$ is the probability of copying word $y^{r}_{t}$ in current position $j$. It is calculated by the revision context $<\textbf{s}_{j},\textbf{c}_{j}>$ of current decoding state and revision contexts of items in the memory.
$\theta_{j}$ is a weight computed at each decoding step:
\begin{align}
\theta_{j}&= \text{sigmoid}(W_{s}\textbf{s}_{j}+W_{c}\textbf{c}_{j}),
\end{align}

If the output word $w$ is an unknown word (\textit{UNK}), but has been corrected by human, our method could successfully generate $w$ in future translation by copying it from the revision memory, which partially alleviate the problem of UNK. To keep the translation process going, the embedding of {{UNK}} is fed into the decoder, as the input for the next time step. 

\begin{table*}[t]
\centering
\begin{tabular}{l|c|c|cc|c|c}
\hline
Model&Revisions &NIST03 & NIST04 & NIST05&Ave.&$\Delta$ \\
\hline
RNNSearch&$-$&37.27&37.25&33.64&35.45&$-$ \\
\hline
\multirow{2}{*}{$\text{UniDiR}$}&1&39.35&39.01&37.22&38.12&+2.67 \\
\cline{2-7}                         
                                    &2&42.09&42.46&40.64&41.55&+6.10\\
\cline{1-7}
\hline
\multirow{2}{*}{$\text{UniDiR}_{\text{G}}$}&1&48.88&47.36&44.25&45.81&+10.36 \\
\cline{2-7}                         
                                   &2&52.58&51.95&47.47&49.71&+14.26\\
\hline
\multirow{2}{*}{BiDiR}&1&49.05&49.22&45.32&47.27&+11.82 \\
\cline{2-7}                         
&2&\textbf{53.19}&\textbf{53.70}&\textbf{49.71}&\textbf{51.71}&\textbf{+16.26}\\
\hline
\end{tabular}
\caption{\label{tab: ideal_result} The comparison of translation qualities between uni-directional and \method's BiDiR decoder on the NIST ZH-EN task. }
\end{table*}

\subsubsection{Online Learning}
Adjacent sentences processed in one \textit{session} are highly related, because they are always in the same discourse or domain.
In this sense, former revised sentences may greatly help the  translation for the following sentences. We use an online learning approach~\cite{Alvaro2017Online,peris2018active} to learn from previously corrected sentences in the same session.

The process starts with the basic model trained on the parallel training data. At the beginning  of each session, the basic model is used for translation.
The parameters of the basic model is fine-tuned after \textit{each round} to maximize the generation probability of the revised sentence at current \textit{round}. 
Formally, every time we obtain a correct translation pair $\{\mathbf{x}, \mathbf{y'}\}$ after interaction, we update the whole translation model for one step, according to Equation \ref{loss}. 
By learning from the sentence level interaction history, our Seq2Seq model better fits sentences of current session.
Especially for sessions containing large amounts of sentences, this continuous learning approach may significantly improve the interactive efficiency. 
\section{Experiment}
\subsection{Setup}
\subsubsection{Data-Set}
We conduct experiments on Chinese-English (ZH-EN) and English-Chinese (EN-ZH) translation tasks. For both ZH-EN and EN-ZH, we use NIST data-set to evaulate the proposed framework. The training data consists of about 1.6 million sentence pairs.\footnote{includes \texttt{LDC2002E18}, \texttt{LDC2003E14}, \texttt{LDC2004T08}, \texttt{LDC2005T06}}
We use \texttt{NIST03} as our validation set, and  {\texttt{NIST04}} and {\texttt{NIST05}} as our test sets. These sets have 919, 1597 and 1082 source sentences, respectively, with 4 references. In EN-ZH, we use ref0 of each data set as source sentences.
We extract about 0.2 million sentence pairs\footnote{includes \texttt{LDC2003E14}}  in our training set, which retains the discourse information for training parameters of the revision memory. 
Furthermore, we also use IWSLT2015 data-set~\cite{Cettolo2012WIT3} on the ZH-EN translation task. We use the dev2010 as our validation set and the tst2010-13 as our test set.
We also sample 100 sentences randomly from test sets for the human evaluation.

\subsubsection{Implementation Details}
We test our proposed approach on both of the RNNSearch~\cite{Bahdanau2015Neural} and Transformer~\cite{vaswani2017attention} baselines.
We train the bi-directional NMT model with the sentences of length up to 50 words. For the RNNSearch, vocabularies of both Chinese and English includes the most frequent 30K words for both Chinese and English. We map all out-of-vocabulary words to the special token \emph{UNK}. The dimension of word embedding is 512, and the size of hidden layers is 1024. We use the gradient descent approach to update the parameters, with a batch size of 80. The learning rate is controlled by Adam~\cite{kingma2014adam}.
For the Transformer, we apply byte pair encoding (BPE)~\cite{sennrich2015neural} to all languages and limit the vocabulary size to 32K. We set the dimension of input and output of all layers as 512, and that of feed-forward layer to 2048. We employ 8 parallel attention heads. The number of layers for the encoder and decoder are 6. 
Other settings is same as~\newcite{vaswani2017attention}.
We use beam search for heuristic decoding, and the size is set as 4. 

We measure the translation quality with the IBM-BLEU score~\cite{Papineni2002bleu}.
We implement our interactive NMT model upon NJUNMT-pytorch\footnote{https://github.com/whr94621/NJUNMT-pytorch}. 
The forward and backward decoders of the bi-directional model are trained together with a shared encoder. The learning rate of online learning is $10^{-5}$.
Following previous work~\cite{Tu2017learning}, we train the revision memory by randomly sampling words and contexts in the same discourse. 
The size of revision memory is 100 and it will be used when revision number is more than 20. 
When the revision number is more than 100, the first part will be discarded until revision number is less than 100.

\subsubsection{Ideal and Real Interactive Environments}
Following previous work~\cite{cheng2016a,Alvaro2016Interactive,hokamp2017lexically}, we experiment on both the ideal and real environments.
Because real-world human interactions are expensive and time-consuming to obtain, we first report results on the ideal environment, which generates simulated human interactions by identifying critical mistakes.
The simulated critical mistake is those lead to the most significant BLEU score improvement after being corrected. 

We also validate the interactive efficiency on the real environment, in which three human annotators are asked to revise the translation, until obtaining correct translations.
The three human annotators are asked to work with all different systems with no idea of which system they are working with.

\begin{table}[t]
\centering
\begin{tabular}{l|c|c|c|c}
\hline
Revisions&1&2&3&4\\
\hline
RNNSearch&\multicolumn{4}{c}{35.45}\\
\hline
BiDiR&47.27&51.71 &55.84&59.91 \\
\hline
\hline
Transformer&\multicolumn{4}{c}{43.65}\\
\hline
BiDiR&50.58&53.66&55.73&57.21\\
\hline
\end{tabular}
\caption{\label{tab: multi_revision} Average BLEU score of our \method's bi-directional decoder after multiple interactions on the NIST ZH-EN task. }
\end{table}
\subsection{Results on the Ideal Environment}
We first evaluate \method for ZH-EN on the ideal interactive environment. \newcite{tan2017neural} demonstrate that two revisions already can lead to very good
translation outputs, which is also verified in our experiments on the real interactive
environment.
We also report the results of our model after multiple interactions as a complement.

\subsubsection{Sequential Bi-directional Decoding} First, we valid our proposed bi-directional interactive model~(BiDiR) in \method on the RNN based NMT model (RNNSearch). 
As shown in Table \ref{tab: ideal_result}, BiDiR significantly outperforms the uni-directional~(UniDiR) counterpart, obtaining more than 10 absolute BLEU score improvements. 

For fair comparison, we also report results of \newcite{hokamp2017lexically}~($\text{UniDiR}_{\text{G}}$), which works in a uni-directional fashion, but using a grid beam search.
This enables the uni-directional decoder to revise the critical mistake first as in our bi-directional decoder, but the generation process is also uni-direction.
The results for their model still fall behind ours in performance, and the gap may increase on the real environment.
Because mistakes left to the revision still exist, and could not be fixed automatically.
This harms their interactive efficiency.


We also apply BiDiR decoder on Transformer with the same setting of RNNSearch.
By revision once, BiDiR achieves an absolute BLEU improvement of 6.93, upon a strong Transformer baseline~(43.65 to 50.58), significantly outperforming UniDiR~(+2.04) and $\text{UniDiR}_{\text{G}}$~(+4.57).
We also compare RNNSearch and Transformer with our proposed bidirectional decoder in Table \ref{tab: multi_revision}.
Results show that although the baseline of Transformer is significantly better than RNNSearch, both of them give pretty high BLEU scores~(around 60) after 4 revisions.
RNNSearch even obtains higher BLEU scores.
For convenience, we will only report results based on RNNSearch in followed experiments.


\begin{table}[t]
\centering
\begin{tabular}{l|cc|c|c}
\hline
Model& NIST04& NIST05&AVE.&$\Delta$ \\
\hline
RNNSearch&37.25&33.64&35.45& $-$ \\
\hline
\ +OL&40.62&37.38&39.00& +3.56 \\
\hline
\ +RM&37.66&34.42&36.04& +0.59 \\
\hline
\tabincell{l}{\method}&\textbf{41.13}&\textbf{37.95}&\textbf{39.54}&\textbf{+4.10} \\
\hline
\hline
BiDiR&53.70&49.71&51.71& $-$ \\
\hline
\ +OL&54.75&49.50&52.13&+0.42 \\
\hline
\ +RM&53.96&49.97&51.97&+0.26 \\
\hline
\tabincell{l}{\method}&\textbf{54.86}&\textbf{50.46}&\textbf{52.66}&\textbf{+0.95} \\
\hline

\end{tabular}
\caption{\label{tab: OL_RM} The performance gain from online learning (OL) and revision memory (RM) on the NIST ZH-EN task. BiDiR is \method without OL and RM .}
\end{table}
\begin{table}[t]
    \centering
    \begin{tabular}{l|c|c|c}
    \hline
    Model&Revisions& BLEU&$\Delta$ \\
    \hline
    RNNSearch&$-$&12.37& $-$ \\
    \hline
    UniDiR&1&14.66& +2.29 \\
    \hline
    CAMIT&1&\textbf{19.49}& \textbf{+7.12} \\
    \hline
    
    \end{tabular}
    \caption{\label{tab: iwslt} The comparison of translation quality between uni-directional interactive NMT and the proposed CAMIT on the IWSLT ZH-EN task.}
    \end{table}

\begin{table}[t]
\centering
\begin{tabular}{m{28pt}|l|m{26pt}|c|c}
\hline
&Model & BLEU&\#Revisions&Costs \\
\hline
\multirow{5}{*}{ZH-EN}&RNNSeach&38.23& $-$& $-$ \\
\cline{2-5}
&$\text{UniDiR}_{\text{G}}$&43.68& 2.76 & 62.92 \\
\cline{2-5}
&\tabincell{l}{\method\\ \ \ -OL-RM}&45.82& 2.19 & 62.34 \\
\cline{2-5}
&\tabincell{l}{\method} &\textbf{46.26}&\textbf{1.81}&\textbf{51.48} \\
\hline
\hline
\multirow{5}{*}{EN-ZH}&RNNSearch&24.52& $-$& $-$ \\
\cline{2-5}
&$\text{UniDiR}_{\text{G}}$&29.55& 2.82 & 36.60 \\
\cline{2-5}
&\tabincell{l}{\method\\ \ \ -OL-RM}&\textbf{30.36} & 2.06 & 30.90 \\
\cline{2-5}
&\tabincell{l}{\method} &30.01&\textbf{1.92}&\textbf{26.88} \\
\hline
\end{tabular}
\caption{\label{tab: real_result} Results on the real environment. \#Revisions: the average revision number. Costs: revising times, seconds per sentence. Note our proposed \method achieves the best results.}
\end{table}
\begin{table}[ht]
\centering
\begin{tabular}{l|c|c}
\hline
Model &Number&$\Delta$\\
\hline
RNNSearch&2052&$-$  \\
\hline
\tabincell{l}{
\ +RM}&1723 & -16.1 \%  \\
\hline
\end{tabular}
\caption{\label{tab: unk_num} \textit{UNK} number after using the revision memory.}
\end{table}
\begin{table*}[t!]
\centering
\begin{tabular}{c|l|c|l}
\hline
\multirow{8}{*}{1}&\multicolumn{2}{l|}{Src.} &
\begin{CJK*}{UTF8}{gbsn}
独立 制片人 应当 协助 维护 言论 自由, 雷福 说
\end{CJK*} \\
\cline{2-4}
&\multicolumn{2}{l|}{Ref.} & independent filmmakers should help to protect freedom of speech, redford said\\
\cline{2-4}
&\multicolumn{2}{l|}{RNNSearch} & \emph{independence and should to uphold} freedom of speech, \emph{UNK}  \\
\cline{2-4}
&\multirow{3}{*}{$\text{UniDiR}_{\text{G}}$} &1&\emph{independence} {\color{orange} \textbf{filmmakers}} {\color{blue}\underline{should help to protect}} freedom of speech, \emph{UNK}\\
\cline{3-4}
&&2&\emph{independence} filmmakers should help to protect freedom of speech, {\color{orange}\textbf{ redford}} {\color{blue}\underline{said}}\\
\cline{3-4}
&&3&{\color{orange}\textbf{independent}} filmmakers should help to protect freedom of speech, redford said\\
\cline{2-4}
&\multirow{2}{*}{\method} &1&{\color{blue}\underline{independent}} {\color{orange}\textbf{filmmakers}} {\color{blue}\underline{should help to protect}} freedom of speech, \emph{UNK}\\
\cline{3-4}
&&2&independent filmmakers should help to protect freedom of speech, {\color{orange}\textbf{ redford}} {\color{blue}\underline{said}}\\
\hline
\hline
\multirow{4}{*}{2}&\multicolumn{2}{l|}{Src.} &
\begin{CJK*}{UTF8}{gbsn}
雷福 呼吁 独立 制片人 协力 防止 言论 自由 遭到 腐蚀
\end{CJK*} \\
\cline{2-4}
&\multicolumn{2}{l|}{Ref.} & redford called on independent filmmakers to help prevent the freedom of speech from being eroded \\
\cline{2-4}
&\multicolumn{2}{l|}{RNNSearch} & \emph{UNK} calls \emph{for independence} to prevent the freedom of speech from being eroded \\
\cline{2-4} 
&\multicolumn{2}{l|}{\tabincell{l}{\method}} & {\color{blue}\underline{redford}} calls \emph{for} {\color{blue}\underline{independent filmmakers}} to prevent the freedom of speech from being eroded \\
\hline
\end{tabular}
\caption{\label{tab: case}Two real cases from the NIST ZH-EN test sets. Words with bold and orange fonts are revised by human. Words with underlines and blue fonts are corrected by the model automatically. Words with Italic fonts are mistakes. }
\end{table*}
\subsubsection{Learning from Interaction History} 
Table \ref{tab: OL_RM} shows that our baseline RNNSearch model obtains a significant BLEU score improvement of 4.1 on the ideal environment by learning from interaction history with both online learning~(+OL) and revision memory~(+RM).
Additionally, we find that incorporating history learning into BiDiR 
can still give about 1 absolute BLEU score improvement~(+0.95), which is achieved on a very strong baseline with over 50 BLEU score points.
This experiment shows that learning from the interaction history is helpful to boost the translation performance.
In such case, human will take less revision number for obtaining satisfactory translations. 
\subsubsection{Effectiveness in Different Domains}
In order to verify the effectiveness of the CAMIT in different domains, we also employ the proposed method on IWSLT ZH-EN translation task. Results are shown in Table \ref{tab: iwslt}, compared with baseline, the proposed CAMIT model could achieve 7.12 BLEU score improvement. Moreover, compared with UniDiR model, the CAMIT also gets a significantly gain.

\subsection{Results on the Real Environment}
We conduct experiments on the real environment for both EN-ZH and ZH-EN on the RNNSearch based system.
We report the average results of 3 volunteers, who could revise arbitrary times until they feel satisfied for the translation.
For better comparison, we include the results of~\newcite{hokamp2017lexically}
in Table \ref{tab: real_result}. Our proposed model \method gives 8.03 and 5.49 BLEU score improvements on ZH-EN and EN-ZH, respectively, with significantly\footnote{P-values are below 0.01 using pairwise t-test.} lower revision numbers than the uni-directional one, and taking less times.

We also compare the number of \emph{UNK} in final translation outputs, between adopting revision memory or not.
Table \ref{tab: unk_num} shows that, the revision memory helps relatively decrease the \emph{UNK} rates by 16.1\%, reducing the number of \emph{UNK} from 2052 to 1723. This indicates that the revision memory enables our model to remember the past revisions, which helps avoid making same mistakes, e.g., \textit{UNK}.

\subsection{Case Study}
We list two real cases of the interactive translation from $\text{UniDiR}_{\text{G}}$~\cite{hokamp2017lexically} and our propsed model in Table \ref{tab: case}.
The two cases are related, and the second case is the next sentence of the first one in a discourse.
For the first one, the baseline NMT system makes several mistakes.
Our proposed bi-directional interactive model corrects these errors by two revisions, while the uni-directional method takes three revisions.
After correcting the word ``filmmakers'', our bi-directional model automatically fixes the mistake of ``\textit{independent}'' left to the revision position while the uni-directional model cannot do this.


The second case shows the translation output after using online learning and revision memory, which could reduce the revision number in practice.
After using the first revised sentence to tune the model by online training, and remembering past revisions by the revision memory, our model have more prior knowledge that the session is related to the topic of \textit{movie}. So the revised word ``redford'' and ``filmmakers'' can be fixed automatically.




\section{Related Work}
{Interactive machine translation has been widely exploited  to improve the translation by using interaction feedback from human users~\cite{langlais2000transtype,simard2007rule,barrachina2009statistical,gonzalez2013interactive,cheng2016a} in statistic machine translation~(SMT)~\cite{Yamada2001,Koehn:2003:SPT:1073445.1073462,Chiang:2007:HPT:1268656.1268659}. 

Recently, researchers employ it in neural machine translation (NMT).
\newcite{barrachina2009statistical}, \newcite{gonzalez2013interactive} and \newcite{knowles2016neural} present an interactive NMT model with the uni-directional interaction protocol (UniDiR), in which users can only interact with the model from left to right. 

\newcite{Alvaro2016Interactive} purpose a new protocol for NMT, which is different to UniDiR, using the average of word probability to make use of the feedback. \newcite{hokamp2017lexically} and \newcite{post2018fast} propose a grid beam search algorithm, in which users can be allowed to make any interaction on arbitrary position with the UniDiR model. 
But their interactive model also regenerate translations from left to right, which omits errors left to the revision. In our proposed model, human can revise the most critical mistake first in arbitrary position of the sentence; and after that the whole sentence will updated, fixing minor mistakes automatically. 

\newcite{Liu2016Agreement} propose the bi-directional decoding model to improve translation quality. 
However, in their approach, two decoders individually generate outputs. 
The translation output of our model is obtained by coupling the forward and backward decoders sequentially. 
\newcite{Alvaro2017Online} propose online learning method to improve the performance of NMT, however, their model is independent of the interactive process. External information, such as word or char, has been verified to be effective in promoting translation~\cite{chen2018combining,zheng2018learning}, but these methods can not employ on the interactive process directly.
Our proposed revision memory is inspired by the \textit{CopyNet}~\cite{gu2016incorporating} and \textit{history cache}~\cite{Tu2017learning}, which
only focus on better using source or global context in supervised learning, and is different from our model.}

\section{Conclusion}
In this paper, we propose \method, to improve the interactive efficiency of NMT models in both decoding and learning aspects. 
For decoding, we propose an sequential bi-directional decoder, updating the whole sentence after each revision.
For learning, we exploit the interaction history for the sentence level and word level.
Experiments show that our model significantly improve the efficiency of interactive NMT.

\section*{Acknowledgements}
We would like to thank the anonymous reviewers for their insightful comments. Thanks to Zewei Sun and Zaixiang Zheng for their insightful comments. Thanks to the volunteers for their hard work on the human evaluation experiment. 
This work is supported by the National Science Foundation of China (No. U1836221, 61672277), the Jiangsu Provincial Research Foundation for Basic Research (No. BK20170074).
\bibliographystyle{named}
\bibliography{ijcai19}

\end{document}